\PassOptionsToPackage{table}{xcolor}
\documentclass[sigconf]{acmart}

\usepackage{subcaption}
\usepackage{multirow}
\AtBeginDocument{%
  }

\copyrightyear{2025}
\acmYear{2025}
\setcopyright{acmlicensed}\acmConference[MM '25]{Proceedings of the 33rd ACM International Conference on Multimedia}{October 27--31, 2025}{Dublin, Ireland}
\acmBooktitle{Proceedings of the 33rd ACM International Conference on Multimedia (MM '25), October 27--31, 2025, Dublin, Ireland}
\acmDOI{10.1145/3746027.3755364}
\acmISBN{979-8-4007-2035-2/2025/10}

\begin{document}

\title{Debiasing Multimodal Large Language Models via Penalization of Language Priors}

\author{YiFan Zhang}
\authornote{Both authors contributed equally to this research.}
\orcid{0000-0002-6227-0183}
\affiliation{%
  \institution{Institute of Automation, Chinese Academy of Sciences}
  \city{Beijing}
  \country{China}
}
\email{yifanzhang.cs@gmail.com}

\author{Yang Shi}
\authornotemark[1]
\affiliation{%
  \institution{Peking University}
  \city{Beijing}
  \country{China}}
\email{frankyang@stu.pku.edu.cn}

\author{Weichen Yu}
\affiliation{%
  \institution{Carnegie Mellon University}
  \city{Pittsburgh}
  \country{United States}
}
\email{wyu3@andrew.cmu.edu}

\author{Qingsong Wen}
\authornote{Corresponding Author.}
\affiliation{%
 \institution{Squirrel AI Learning}
 \city{Bellevue}
 \country{United States}}
\email{qingsongedu@gmail.com}

\author{Xue Wang}
\affiliation{%
 \institution{Alibaba Group}
 \city{Beijing}
 \country{China}}
\email{wxie91@gmail.com}

\author{Wenjing Yang}
\affiliation{%
 \institution{National University of Defense Technology}
 \city{Changsha}
 \country{China}}
\email{wenjing.yang@nudt.edu.cn}

\author{Zhang Zhang}
\affiliation{%
 \institution{Institute of Automation, Chinese Academy of Sciences}
 \city{Beijing}
 \country{China}}
\email{zzhang@nlpr.ia.ac.cn}

\author{Liang Wang}
\affiliation{%
 \institution{Institute of Automation, Chinese Academy of Sciences}
 \city{Beijing}
 \country{China}}
\email{wangliang@nlpr.ia.ac.cn}

\author{Rong Jin}
\affiliation{%
 \institution{Meta}
 \city{Seattle}
 \country{United States}}
\email{rongjinemail@gmail.com}

\renewcommand{\shortauthors}{YiFan Zhang et al.}

\begin{abstract}
In the realms of computer vision and natural language processing, Multimodal Large Language Models (MLLMs) have become indispensable tools, proficient in generating textual responses based on visual inputs. Despite their advancements, our investigation reveals a noteworthy bias: the generated content is often driven more by the inherent priors of the underlying Large Language Models (LLMs) than by the input image. Empirical experiments underscore the persistence of this bias, as MLLMs often provide confident answers even in the absence of relevant images or given incongruent visual inputs. To rectify these biases and redirect the model's focus toward visual information, we propose two simple, training-free strategies. First, for tasks such as classification or multi-choice question answering, we introduce a ``\textbf{Post-Hoc Debias}'' method using an affine calibration step to adjust the output distribution. This approach ensures uniform answer scores when the image is absent, acting as an effective regularization technique to alleviate the influence of LLM priors. For more intricate open-ended generation tasks, we extend this method to ``\textbf{Visual Debias Decoding}'', which mitigates bias by contrasting token log-probabilities conditioned on a correct image versus a meaningless one. Additionally, our investigation sheds light on the instability of MLLMs across various decoding configurations. Through systematic exploration of different settings, we achieve significant performance improvements—surpassing previously reported results—and raise concerns about the fairness of current evaluation practices. Comprehensive experiments substantiate the effectiveness of our proposed strategies in mitigating biases. These strategies not only prove beneficial in minimizing hallucinations but also contribute to the generation of more helpful and precise illustrations.
\end{abstract}





\maketitle

\section{Introduction}
Large Language Models (LLMs) such as GPT~\cite{gpt4}, LLaMA~\cite{dubey2024llama}, and Qwen~\cite{yang2024qwen2} have been extended to handle both textual and visual inputs. This expansion into the multimodal domain typically involves additional pretraining on image-text pairs, followed by fine-tuning with specialized visual instruction tuning datasets. These steps have led to the emergence of powerful Multimodal Large Language Models (MLLMs)~\cite{liu2023improvedllava, qwen2vl, fu2025vita, shi2025mavors, liu2024nvila,chen2025versavid}. Capable of processing complex visual inputs, MLLMs can perform visual understanding and reasoning tasks, enabling a broad range of applications, including document processing~\cite{shi2025mme}, healthcare~\cite{li2024llava, qiu2024application}, embodied agent~\cite{chen2023towards, cheng2025embodiedeval}, and autonomous driving~\cite{cui2024survey, chen2024advanced}.

\begin{figure}[t]
    \centering
    \includegraphics[width=\linewidth]{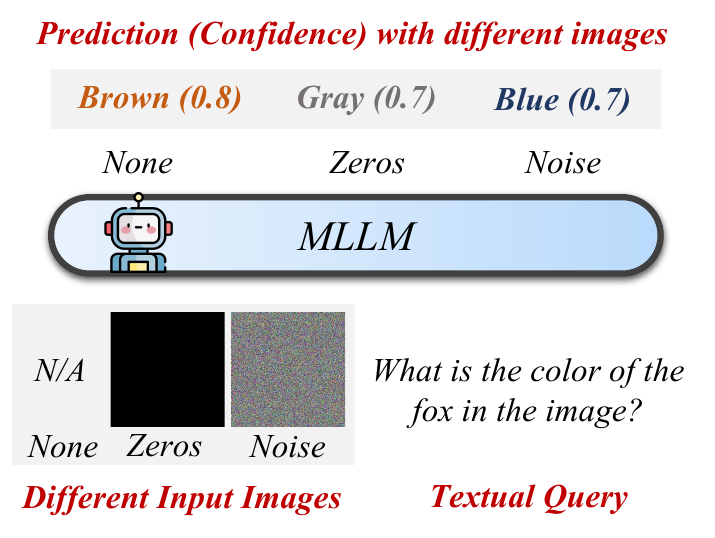}
    \vspace{-2em}
    \caption{MLLM tends to generate confident answers even when presented with nonsensical or irrelevant images, thereby revealing a pronounced bias towards the learned language patterns. ``None'' indicates the absence of an input image, while ``Noise'' signifies the presence of Gaussian noise matching the image dimensions. ``Zeros'' indicates the input is a tensor composed entirely of zero values.}
    \label{fig:spurr_color}
\end{figure}

Despite the impressive overall performance of MLLMs, our investigation uncovers a critical limitation in their reliance on visual input. Specifically, their outputs tend to reflect patterns learned during language pretraining, often at the expense of proper grounding in visual data. Even when presented with entirely noisy or absent images, MLLMs consistently generate responses with unwarranted confidence, indicating a strong bias towards the learned language patterns. This observation is illustrated in Figure~\ref{fig:spurr_color}, where LLaVA-1.5~\cite{liu2023improvedllava} generates confident but incorrect answers—such as brown, gray, or blue—when asked about the color of a non-existent fox.

\begin{figure}[htp]
    \centering
    \includegraphics[width=\linewidth]{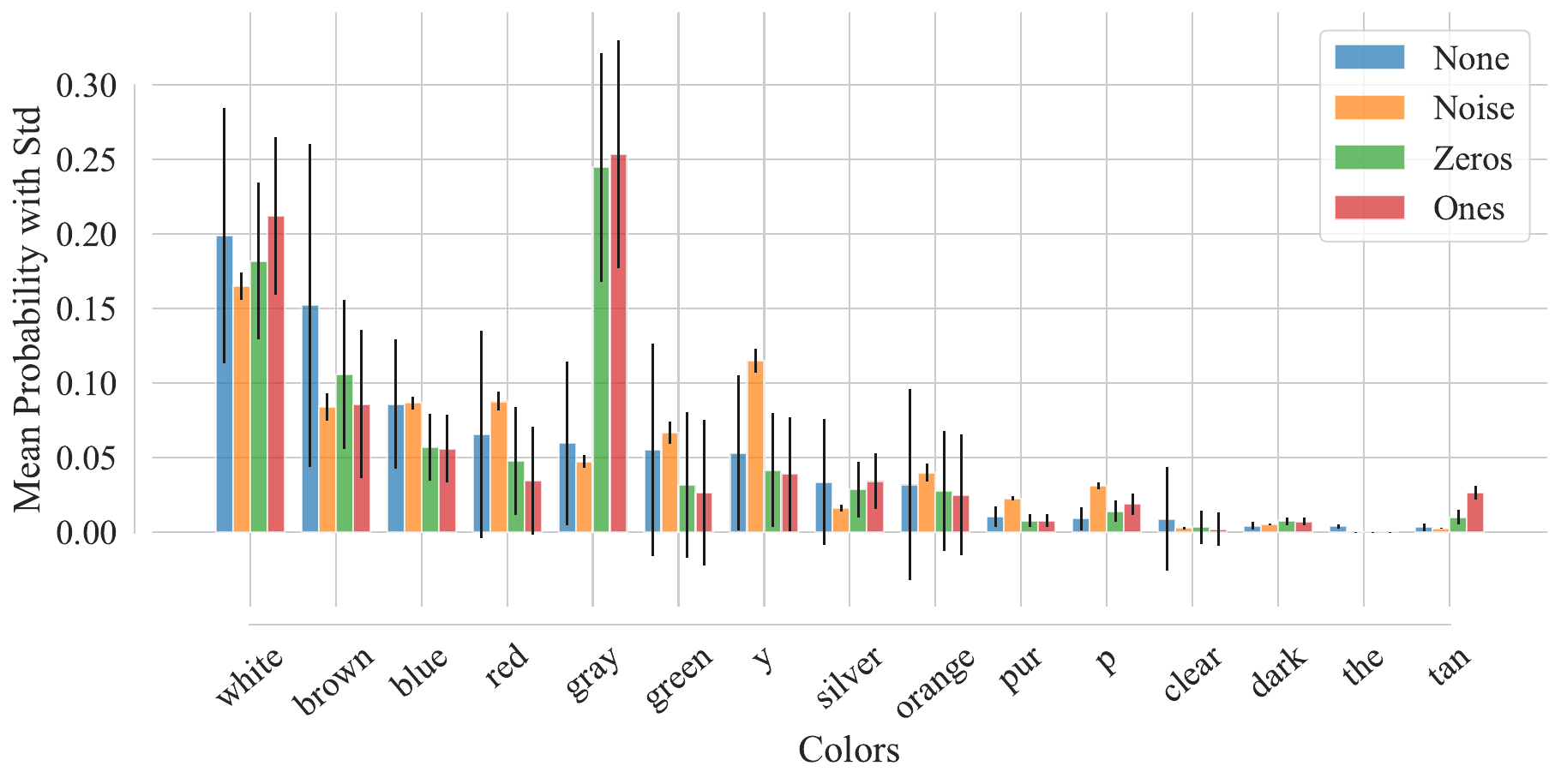}
    \vspace{-2em}
  \caption{Top 15 answer choices and their probabilities.}
  \label{fig:toy_experiment}
\end{figure}

To further investigate this bias, we conduct a toy experiment involving questions about color, shape, number, actions, and relationships for 80 entities from the MSCOCO dataset~\cite{lin2014microsoft}. We evaluate model behavior under several baseline conditions: \textit{Noise} (a fully noisy image with a corresponding question), \textit{None} (no image input), and \textit{Zeros/Ones} (images filled with all-zero or all-one values). Figure~\ref{fig:toy_experiment} displays the top 15 answer choices and their corresponding probabilities when the model is asked about the color. Remarkably, even when no image is provided or when questions reference non-existent or meaningless visual content, current MLLMs still tend to generate specific responses. This behavior highlights a strong bias rooted in language model pretraining. Such biases present persistent challenges—particularly hallucination~\cite{gunjal2023detecting, li2023evaluating, sun2023aligning}—and raise serious concerns about the reliability and applicability of MLLMs.

\begin{figure*}[t]
  \centering
  \includegraphics[width=\linewidth]{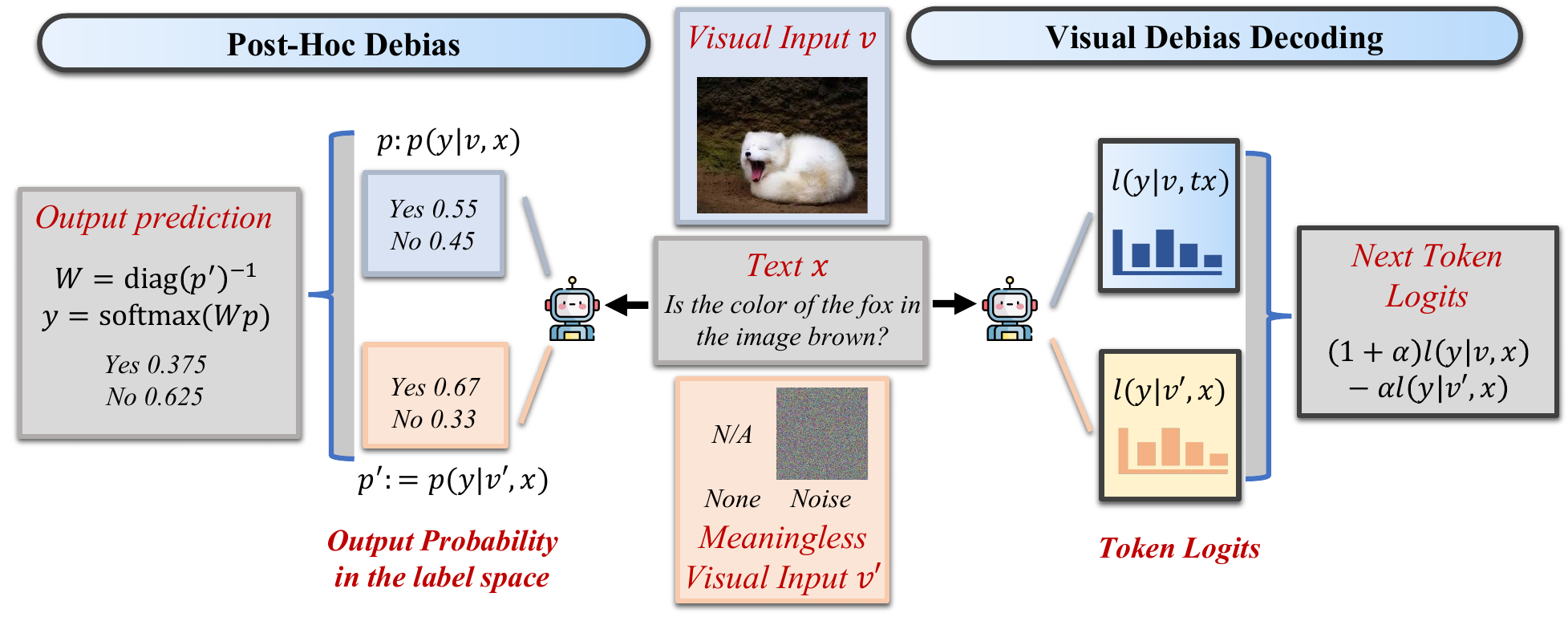}
  \caption{Illustration of the proposed Post-Hoc Debias and Visual Debias Decoding methods. The former focuses on debiasing the prediction results, while the latter modifies the next-token distribution.}
  \label{fig:debias_model}
\end{figure*}

In this paper, we aim to uncover the underlying causes of the phenomenon and offer insights into the design and training strategies of current MLLMs. Building on this analysis, we present two training-free strategies to mitigate biases inherent in the underlying LLMs, which are depicted in Figure~\ref{fig:debias_model}. Our first approach introduces a ``calibration'' step to adjust the output distribution. Specifically, we assess the model's inclination toward certain answers by providing a dummy input without an image—replacing the image with either None or meaningless tensors, as illustrated in Figure~\ref{fig:spurr_color}. We then determine the calibration parameters to ensure that the image-free input yields uniform scores across possible answers, avoiding overly confident outputs. This procedure, referred to as the ``\textbf{Post-Hoc Debias}'' approach, establishes appropriate parameters without requiring additional training data and is particularly effective for classification tasks. However, in open-ended generation tasks, calibrating each token's logits uniformly, even without image input, is unreasonable due to the inherent correlation between text tokens. To address this, we extend the Post-Hoc Debias method to its ``\textbf{Visual Debias Decoding}'' counterpart. Inspired by contrastive decoding methods~\cite{li2022contrastive,leng2023mitigating}, this approach computes the difference in token-level log-probabilities between inputs containing a meaningful image and those with a meaningless visual input. Both the Post-Hoc Debias and Visual Debias Decoding methods act as regularization techniques, reducing the model’s reliance on textual prompts or meaningless visual inputs during generation.

\begin{figure}[htp]
  \centering
  \includegraphics[width=\linewidth]{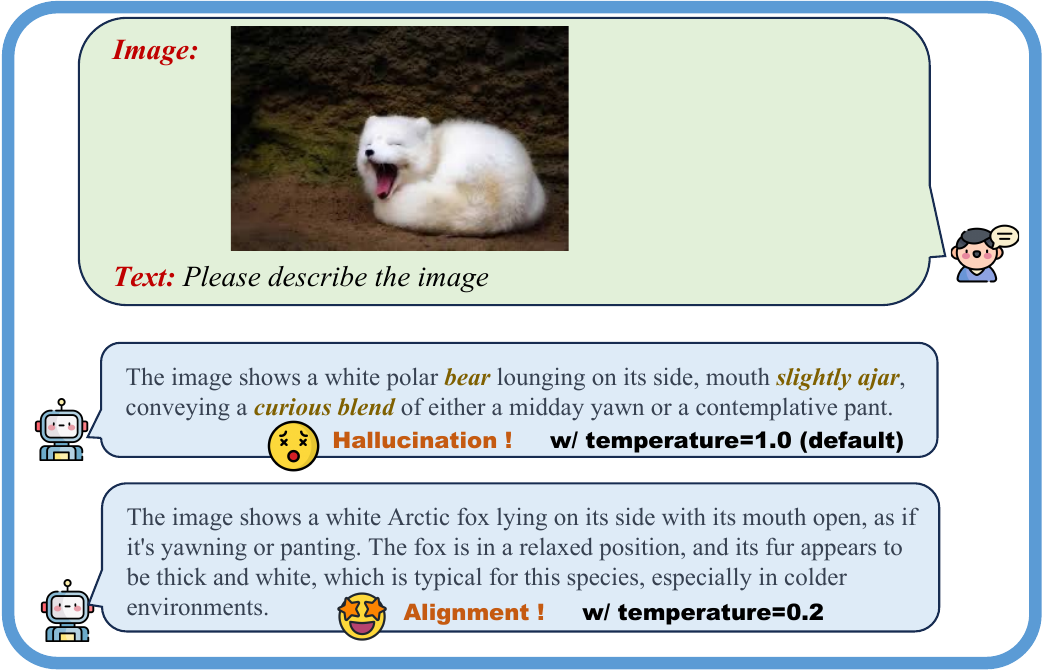}
  \vspace{-2em}
  \caption{A simple adjustment of the temperature from 1.0 to 0.2 results in the
  successful generation.}
  \label{fig:sampling_teaser}
\end{figure}

Moreover, we observe substantial performance instability in current MLLMs across different generative configurations. Our primary hypothesis is that existing evaluations are predominantly based on a default decoding setting, limiting the exploration of the model's full capabilities. Our investigation reveals that different models favor distinct generation configurations, as illustrated in Figure~\ref{fig:sampling_teaser}, underscoring the instability of current MLLM evaluations. This raises concerns about evaluation fairness, especially given the common reliance on default generation settings or selectively optimized configurations tailored to the proposed model. To fully realize the potential of existing MLLMs, we conduct a systematic evaluation of six models across four benchmarks, thoroughly exploring their generative configurations to identify optimal settings. Our results significantly outperform previously reported ones by a substantial margin, emphasizing the critical role of configuration selection in fair and accurate model evaluation.

\section{Related Work}
Hallucination is redefined from the natural language processing community's original concept of generating factual yet ungrounded content to encompass content not supported by the associated image~\cite{tonmoy2024comprehensive,ji2023survey}. Several strategies have been explored to mitigate hallucinations in MLLMs. Initial efforts, geared towards small-scale MLLMs, involved fine-grained modality alignment~\cite{rohrbach2018object} and tackling statistical bias in object co-occurrence through data augmentation~\cite{kim2023exposing}. However, the distinctive behaviors of MLLMs pose a significant challenge, rendering existing methods impractical for generalization and scaling up~\cite{wei2022emergent}. Recent studies have addressed this challenge by introducing hallucination-targeted datasets for fine-tuning~\cite{gunjal2023detecting,liu2023aligning}, training a post-hoc revisor to generate less hallucinatory outputs~\cite{zhou2023analyzing}, or adopting factually augmented Reinforcement Learning from Human Feedback (RLHF)~\cite{sun2023aligning, mm_rlhf}. While these interventions effectively address object hallucination in MLLMs, the associated human effort and computational costs underscore the need for a simpler yet efficient approach. A closely related method, Visual Contrastive Decoding (VCD)~\cite{leng2023mitigating}, manipulates the decoding distribution to reduce hallucination. To this end, we introduce two approaches that strike a balance between efficiency and effectiveness.


\vspace{-15pt}

\section{Preliminaries}

\noindent\textbf{Probing Language Bias in MLLMs.} According to the related work, despite the acknowledged influence of language priors on VQA tasks, previous efforts have primarily concentrated on mitigating biases within the training sets. These approaches involve designing advanced training strategies and constructing balanced datasets. However, extending such methodologies to MLLMs, which rely on pretrained LLMs for language understanding and generation, is challenging due to the complexity and computational cost of retraining or fine-tuning. Furthermore, it remains unclear whether MLLM performance is affected by the language priors ingrained in their used LLMs.

\begin{figure*}[htbp]
  \centering
  \begin{subfigure}[t]{0.3\linewidth}
    \centering
    \includegraphics[width=\linewidth]{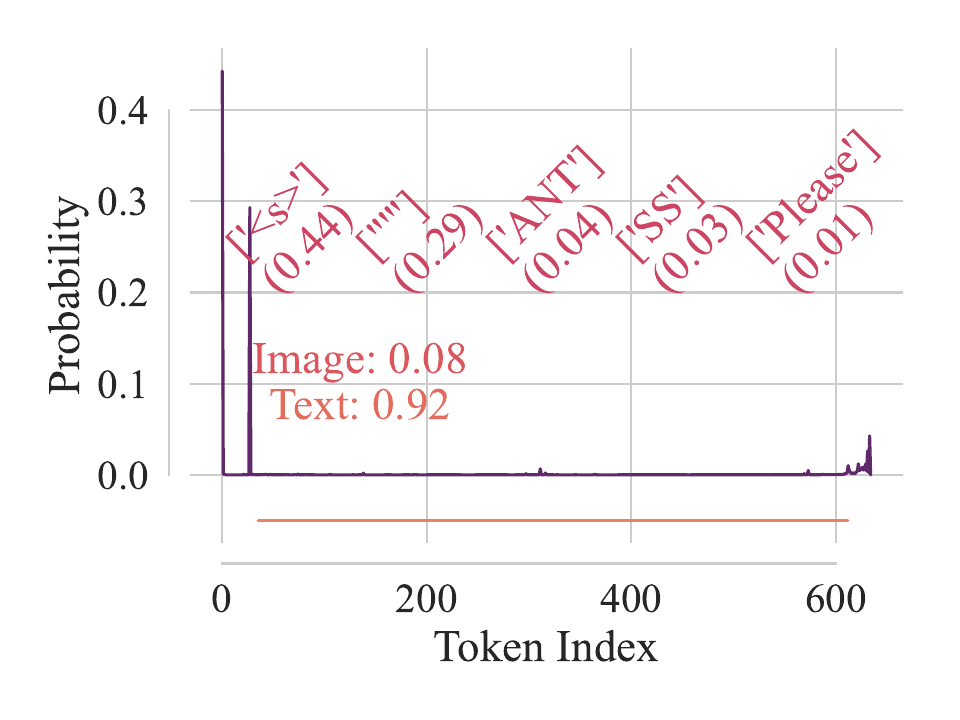}
    \vspace{-2em}
    \caption{Naive}
    \label{30_naive}
  \end{subfigure}
  \hfill
  \begin{subfigure}[t]{0.3\linewidth}
    \centering
    \includegraphics[width=\linewidth]{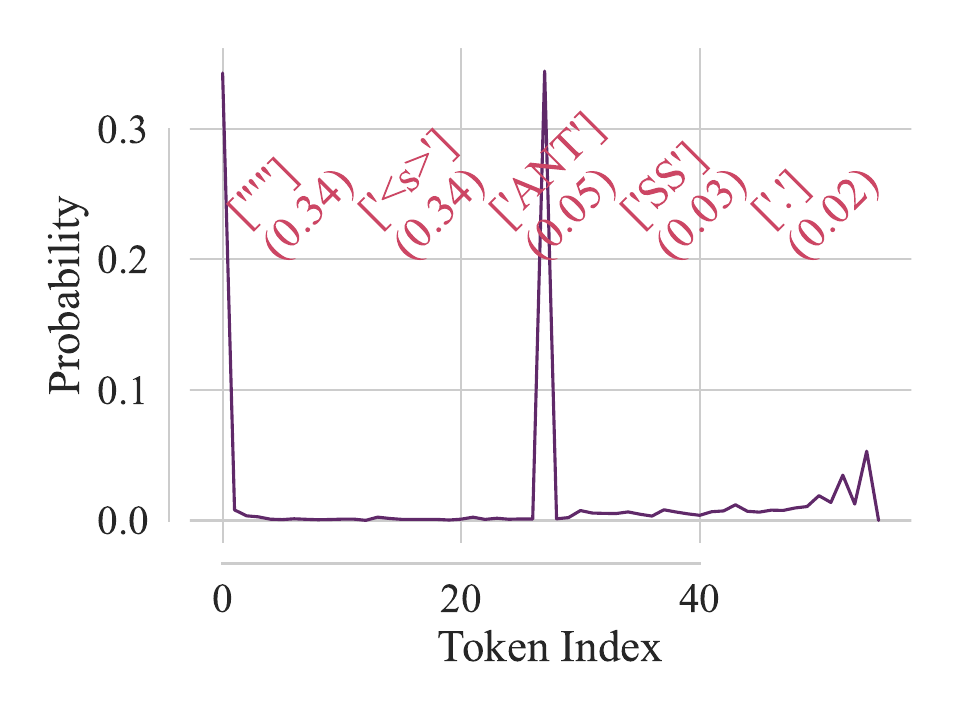}
    \vspace{-2em}
    \caption{None}
    \label{30_none}
  \end{subfigure}
  \hfill
  \begin{subfigure}[t]{0.3\linewidth}
    \centering
    \includegraphics[width=\linewidth]{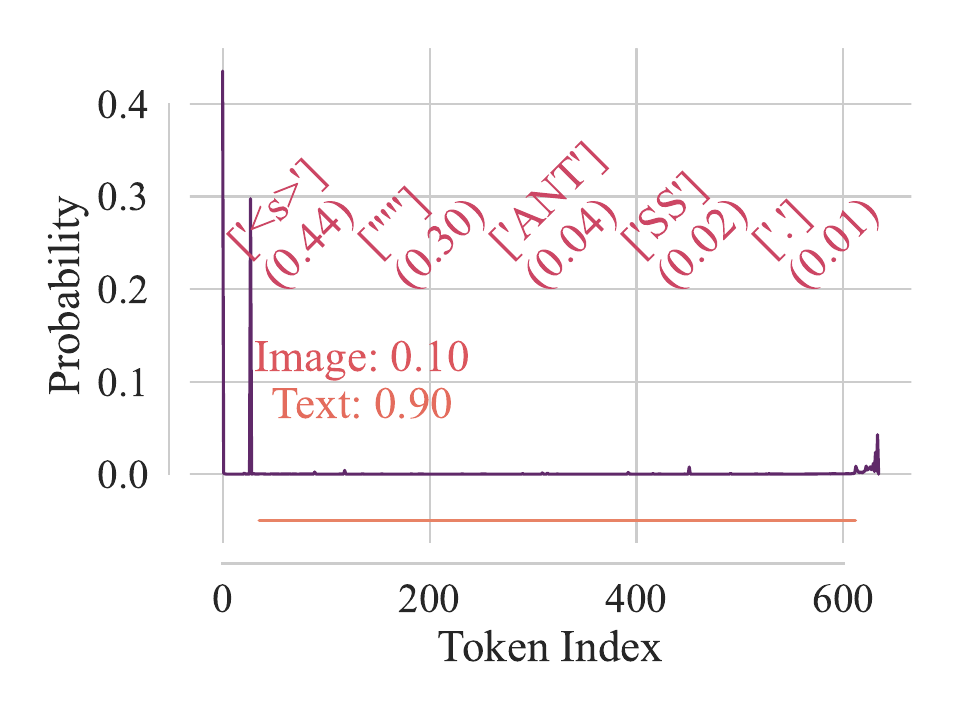}
    \vspace{-2em}
    \caption{Noise}
    \label{30_noise}
  \end{subfigure}
  \vspace{-1em}
  \caption{Attention maps for different input images. The question is ``\textit{A chat between a curious human and an artificial intelligence assistant. The assistant gives helpful, detailed, and polite answers to the human's questions. USER: $<image>$ Is there a bed in the image? Please answer this question with one word. ASSISTANT:}''.}
  \label{fig:atten_calibrate_30}
\end{figure*}


To assess the influence of LLM biases on MLLMs, we conduct toy experiments with five target questions about color (e.g., What is the color of the cat in the image?), shape (e.g., What is the shape of the cat in the image?), and so on. For each entity in the MSCOCO~\cite{lin2014microsoft}, we construct corresponding prompts and pair them with different types of visual inputs. These include unrelated visual inputs such as fully noisy images (Noise) or completely black or white images (Zero/One). We also explore text-only settings by either removing all visual tokens (None) or replacing them with meaningless placeholders such as $<unk>$ (Unk). As illustrated in Figure~\ref{fig:toy_experiment}, MLLMs exhibit biases towards specific answers, such as associating ``white'' with color-related questions, ``round'' with shape-related questions, and ``many'' with number-based questions.

We further analyze the attention mechanisms to investigate why models rely heavily on LLM-induced biases. As is shown in Figure~\ref{fig:atten_calibrate_30}, the outputs tend to allocate disproportionately more attention to text tokens. Even for original image-text pairs in which image tokens outnumber text tokens, over $90\%$ of the total attention is still allocated to text tokens. An intriguing observation is that when the model is fed with the original image-text pair (Naive), pure text (None), or fully noisy images (Noise), it exhibits similar attention patterns. Most of the attention is directed toward special yet uninformative tokens, rather than meaningful visual tokens or relevant text content such as the question itself.

\begin{figure}[t]
  \centering
  \begin{subfigure}[t]{0.48\linewidth}
    \centering
    \includegraphics[width=\linewidth]{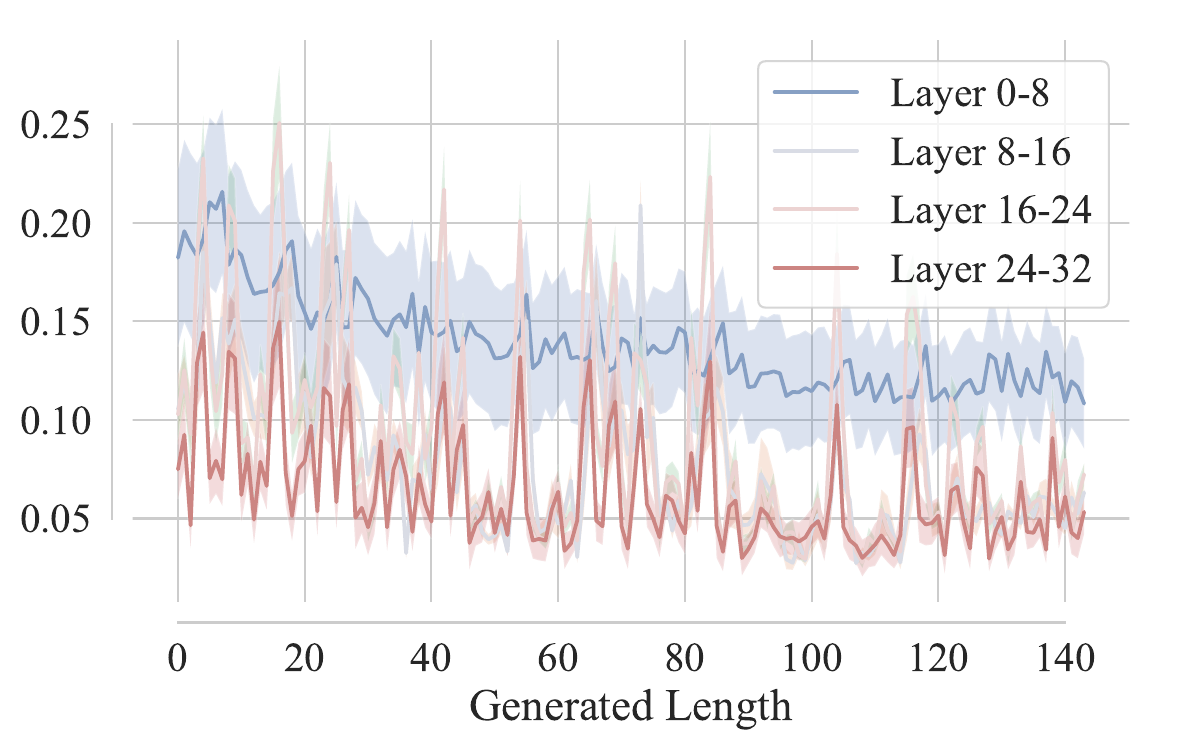}
    \vspace{-2em}
    \caption{Image tokens}
    \label{fig:ImageTokens}
  \end{subfigure}
  \hfill
  \begin{subfigure}[t]{0.48\linewidth}
    \centering
    \includegraphics[width=\linewidth]{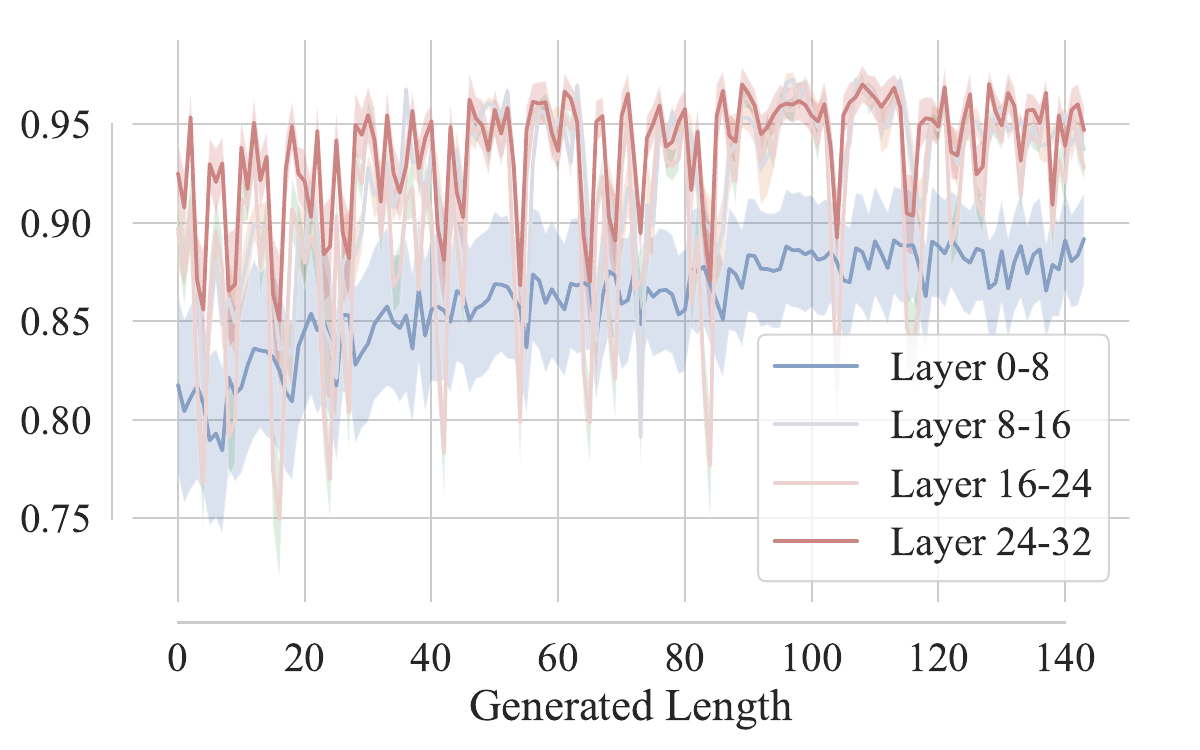}
    \vspace{-2em}
    \caption{Text tokens}
    \label{fig:TextTokens}
  \end{subfigure}
  \vspace{-1em}
  \caption{Attention scores attained by each layer during generation in Sample 1.}
  \label{app:attention_analysis}
\end{figure}
\begin{figure}[t]
  \centering
  \begin{subfigure}[t]{0.48\linewidth}
    \centering
    \includegraphics[width=\linewidth]{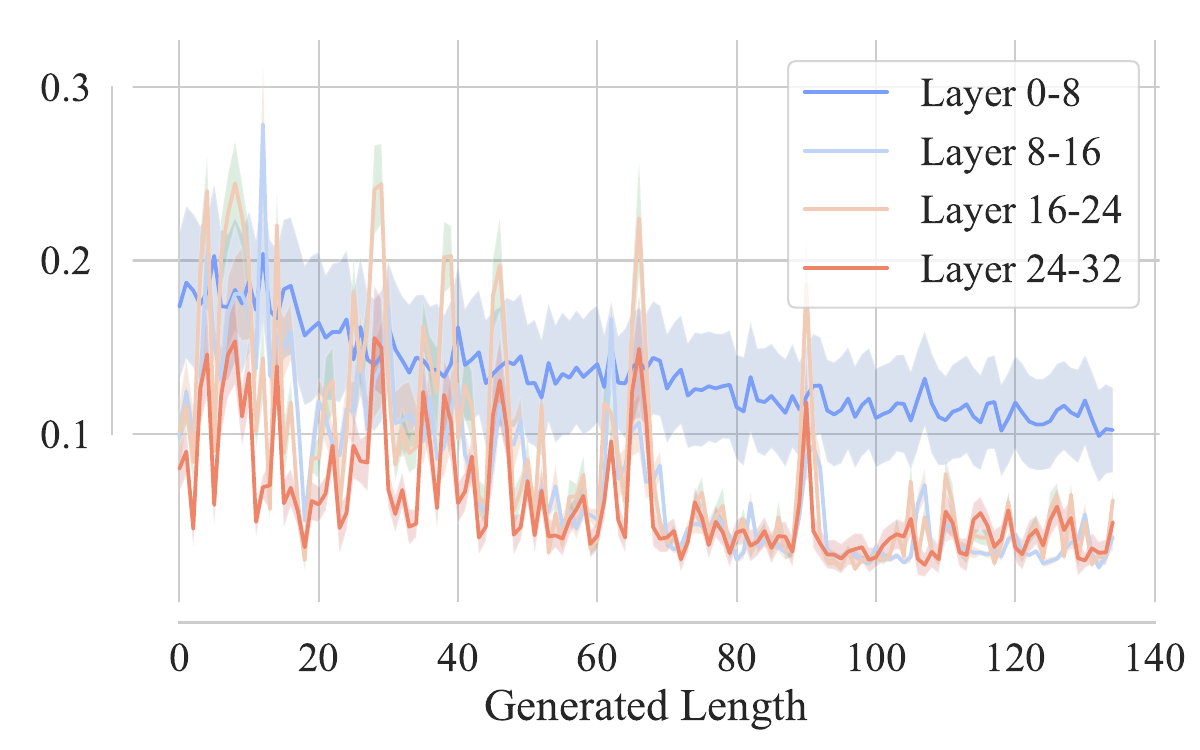}
    \vspace{-2em}
    \caption{Image tokens}
    \label{fig:ImageTokens2}
  \end{subfigure}
  \hfill
  \begin{subfigure}[t]{0.48\linewidth}
    \centering
    \includegraphics[width=\linewidth]{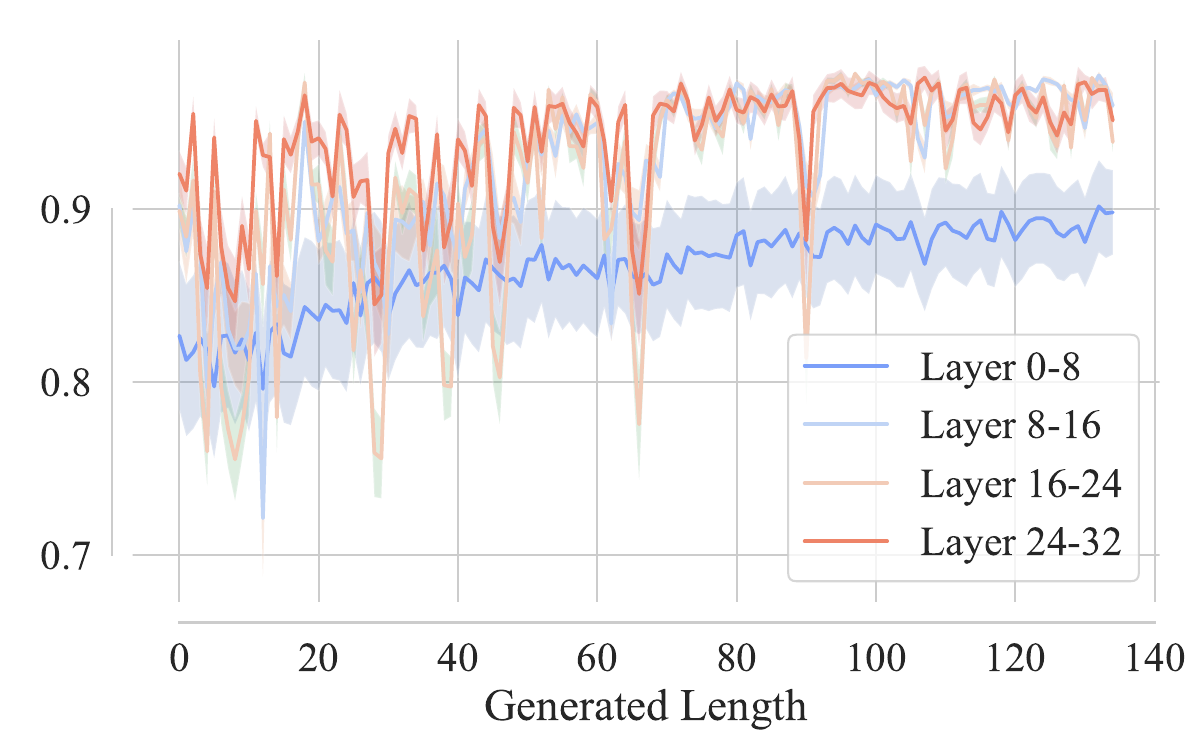}
    \vspace{-2em}
    \caption{Text tokens}
    \label{fig:TextTokens2}
  \end{subfigure}
  \vspace{-1em}
  \caption{Attention scores attained by each layer during generation in Sample 2.}
  \label{app:attention_analysis2}
\end{figure}

In open-ended generation tasks, we randomly select two questions (``Analyze the image in a comprehensive and detailed manner.'' and ``Describe the following image.'') from LLaVA-Bench~\cite{liu2023visual} and examine the attention scores as more tokens are generated. As depicted in Figure~\ref{app:attention_analysis} and Figure~\ref{app:attention_analysis2}, with the increasing length of generated text during open-ended tasks, there was a corresponding reduction in attention allocated to images, exacerbating the issue. Consequently, the model becomes increasingly prone to generating content independently of input images, shedding light on the genesis of hallucinated content from an attention perspective. Additionally, the results also indicate that the shallow layers assign higher attention to visual tokens; however, as the feature progresses into deeper layers, attention to visual tokens will be reduced.

\hfill
  
\noindent\textbf{Underlying Reasons.}
Several factors contribute to this phenomenon, with the modality gap~\cite{NEURIPS2022_702f4db7,wang2022estimating} emerging as a key factor. This gap becomes apparent as image-text embeddings consistently reside in distinct spaces with substantial distances, a characteristic observed even in models trained on a significant number of image-text pairs~\cite{radford2021learning}. Unfortunately, MLLMs consistently undergo training with a restricted vision-language dataset~\cite{sun2023aligning}, resulting in a weakened feature alignment. In contrast, the LLMs involved in the process benefit from a more extensive training corpus and demonstrate heightened proficiency in addressing text-based tasks. The pronounced modality gap guides LLMs to predominantly focus on familiar tokens or representations, often neglecting crucial vision tokens. Additionally, following the shallow-to-deep principle~\cite{sahoo2017online,phuong2019distillation}, where shallower networks can adeptly adapt to changes in data streams or learn more efficiently with limited data, we can explain why shallow layers exhibit better attention score assignments compared to deeper layers. Shallow layers may be better tuned, but deeper layers retain most of the original LLM patterns and tend to ignore unfamiliar vision tokens. Furthermore, MLLMs may inherit the drawbacks of LLMs, which often overlook contextual information and heavily rely on their prior parametric knowledge\cite{shi2023trusting,chen-etal-2022-rich,zhou2023contextfaithful}, further diminishing the impact of vision tokens. These findings highlight inherent limitations in current training strategies and datasets, underscoring the need for improved alignment between modalities or a greater emphasis on the role of visual tokens. To address this issue, we propose two training-free methods aimed at mitigating the observed phenomenon and enhancing the overall performance of MLLMs.

\section{Methodology}

\subsection{Post-Hoc Debias}
As shown in Figure~\ref{fig:debias_model} (left), the \textbf{Post-Hoc Debias} method is specifically tailored for classification tasks or multiple-choice settings, where a predefined set of candidate options is available. In addressing the bias issue, we implement a ``calibration'' step for the model’s output probabilities using an affine transformation~\cite{platt1999probabilistic,guo2017calibration,zhao2021calibrate}:
\begin{equation}
  y=softmax(Wp_\theta+b)
\end{equation}
A weight matrix $W$ and a bias vector $b$ are applied to the original probability $p$ to obtain calibrated probabilities. In classification tasks, $p$ denotes the normalized probabilities over candidate labels, summing to one. To improve computational efficiency, $W$ is constrained to be diagonal, following the vector scaling approach~\cite{guo2017calibration,zhao2021calibrate}. A natural way to estimate $W$ involves capturing the LLM’s inherent bias by giving a meaningless visual input $v’$, such as a noisy image. The resulting prediction is denoted as:
\begin{equation}
  p':=p'_\theta(y|x,v')
\end{equation}
Ideally, the LLM should assign a uniform probability distribution to such inputs. For example, given the question \textit{``Is the color of the fox in the image brown?''}, supplying a meaningless or empty image should yield equal probabilities for ``yes'' and ``no'', as the model lacks sufficient information to decide. However, due to inherent biases, the model often favors certain responses—typically assigning a higher score to ``yes''. This bias can be mitigated by setting:
\begin{equation}
  W=\text{diag}(p'_\theta)^{-1}
\end{equation}
where $p’\theta$ denotes the model’s output under a meaningless visual input, and setting the bias term $b$ to a zero vector. The debiased prediction is then computed as 
\begin{equation}
  y=softmax(Wp_\theta+b)
\end{equation}
Our experimental results show that using unrelated visual inputs (e.g., Noise, Zero, One) does not consistently improve model performance and is sensitive to the type of question. Therefore, we adopt an image-free approach as the default debiasing method, specifically using None and Unk. We also propose a variant that combines both image-free inputs. In this case, we set $v’$ to None and $<$unk$>$, and obtain the corresponding output distributions $p’_n$ and $p’_u$. The final debiasing distribution is then computed as:
\begin{equation}
  p' = (p'_n + p'_u)/2
\end{equation}

\subsection{Visual Debias Decoding}
While Post-Hoc Debias methods can, in principle, be extended to open-ended generation tasks—where the label space includes the entire vocabulary—their implementation is hindered by the large output space. Moreover, it is essential to preserve the semantic coherence and co-occurrence of generated tokens; enforcing a uniform output distribution from the language model may severely degrade generation quality. To address these challenges, we adopt the concept of contrastive decoding~\cite{li2022contrastive,leng2023mitigating,yuan2023speculative,shi2023trusting} and propose the \textbf{Visual Debias Decoding (VDD)} strategy. Following a similar intuition to Post-Hoc Debias, we first obtain logits $l:= l_\theta(y \mid x, v)$ by feeding the original image and text into the MLLM, and then compute contrastive logits $l':= l_\theta(y \mid x, v’)$ using an image-free input, where $l’$ reflects purely text-based priors. To highlight the contribution of visual information, our goal is to suppress undesirable biases captured by $l’$, while preserving the positive, image-conditioned behaviors of the MLLM during generation. To implement this concept, we propose the contrastive objective: 
\begin{equation}
  p_{vdd}(y|v,v',x)=\text{softmax}\left[(1+\alpha) l_\theta(y|x,v)-\alpha l_\theta(y|x,v')\right]
\end{equation}
In this framework, $\alpha$ denotes the amplification level of the debiasing method. It is important to acknowledge that the logits produced by pure-text models are not inherently incorrect; in fact, they often capture essential aspects of English grammar and commonsense reasoning. Consequently, applying a uniform penalty across all tokens may not be appropriate. To address this nuanced issue, we introduce an adaptive plausibility constraint over the output vocabulary $\mathcal{V}$ of MLLMs. This constraint dynamically adjusts based on the model’s confidence in its predictions when visual input is available, and is defined as: 
\begin{equation}
\mathcal{V}_{head}(y_{t})=\{y_i\in\mathcal{V}:p_\theta(y_i|v,t,y_{<t})\geq \beta\max_{w\in\mathcal{V}}p_\theta(w|v,t,y_{<t} )\}.
\end{equation}
Here, $\beta\in[0,1]$ is a hyperparameter that controls the truncation level of the predicted token distribution. A larger $\beta$ results in more aggressive filtering, retaining only the most probable candidates. During generation, we set the logits of all tokens not in $\mathcal{V}_{head}$ to $-\infty$:
\begin{equation}
  p_{vdd}(y_t) = 
  \begin{cases} 
  \begin{aligned}
  \text{softmax}\big[ &(1+\alpha) l_\theta(y|x,v) \\
                     &{}- \alpha l_\theta(y|x,v')\big],
  \end{aligned}
  & \text{if } y_t \in \mathcal{V}_{head}; \\
  0, & \text{otherwise.}
  \end{cases}
  \end{equation}
  The adaptive plausibility constraint selectively retains high-probability tokens based on visual input, while mitigating bias by leveraging pure-text logits. When the model exhibits high confidence in a specific token, the constraint may retain as few as one candidate, thereby diminishing the influence of the contrastive objective. We categorize our methods as VDD-None, VDD-Unk, and VDD-Both, corresponding to the use of image-free inputs—None, Unk, or both—for computing reference logits.

\subsection{Impact of Decoding Configuration}
We observe that existing open-source MLLMs undergo evaluations using a variety of generation strategies. This variability in evaluation configuration introduces a potential source of performance variability, impacting the thorough exploration of MLLM capabilities. Our research primarily delves into decoding configurations, specifically with distinct strategies:

\noindent1. \textbf{Temperature Sampling}: We systematically vary the temperature parameter ($\tau$) from 0.05 to 1, with a step size of 0.05, resulting in 20 configurations. The temperature parameter modulates the sharpness of the next-token distribution.

\noindent2. \textbf{Top-$k$ Sampling}: By filtering the K most likely next words, where K varies within {1, 2, 5, 10, 20, 50, 100, 200, 500}, we generate 9 configurations.

\noindent3. \textbf{Top-$p$ Sampling}: We select words from the smallest set whose cumulative probability exceeds $p$. The threshold $p$ ranges from 0.05 to 1, with a step size of 0.05, yielding 20 configurations.

For each test sample of a given benchmark, the model generates 49 responses, each corresponding to a distinct decoding configuration.. We define four evaluation settings: \textbf{Temp $\tau$}, \textbf{Top-$p$}, and \textbf{Top-$k$}, where the best response is selected from the configurations within each respective group. The \textbf{Overall} setting selects the best response across all 49 configurations. This systematic exploration enables a more comprehensive assessment of MLLM performance under varying decoding strategies. Note that the impact of different sampling strategies is evaluated independently. Unless otherwise specified, all baseline results are reported under the same decoding strategy.

\section{Experiments}
\subsection{Experimental Setup}
This paper adopts a comprehensive evaluation strategy by leveraging four diverse datasets to assess the capabilities of MLLMs. The first dataset, MMMU~\cite{yue2023mmmu}, spans six core disciplines and evaluates models on multimodal questions sourced from college exams, quizzes, and textbooks, with an emphasis on advanced perception and reasoning~\cite{yue2023mmmu}. The second, MME~\cite{fu2023mme}, systematically assesses models’ perceptual and cognitive capabilities through tasks such as OCR, object recognition, and multimodal reasoning~\cite{fu2023mme}. Its subsets, MME-OH and MME-AH, are specifically designed to evaluate object-level and attribute-level hallucinations. Additionally, MME-RE includes reasoning subsets—such as code reasoning, text translation, commonsense reasoning, and numerical computation—to further evaluate cognitive capabilities. The third dataset, POPE~\cite{li2023evaluating}, offers a more refined benchmark for assessing object hallucination in MLLMs. It emphasizes balanced sampling strategies and incorporates data from MSCOCO~\cite{lin2014microsoft}, A-OKVQA~\cite{schwenk2022okvqa}, and GQA~\cite{hudson2019gqa}. Lastly, LLaVA-Bench~\cite{liu2023visual} evaluates the adaptability of MLLMs to challenging tasks and novel domains using 24 diverse images and 60 carefully designed questions, testing the models’ robustness across a range of prompts.

We evaluate the effectiveness of the proposed method on three representative MLLMs: LLaVA-1.5~\cite{liu2023improvedllava} and InstructBLIP~\cite{dai2023instructblip}—both built with Vicuna-7B~\cite{chiang2023vicuna} as the language decoder~\cite{dai2023instructblip, liu2023improvedllava}—as well as Qwen-VL and Qwen-VL-Chat~\cite{bai2023qwen}, based on the Qwen-7B backbone~\cite{qwen1}. To provide a strong baseline, we additionally include two Supervised Fine-Tuning (SFT) models. The first, LLaVA-SFT~\cite{sun2023aligning}, is trained on a novel data mixture that combines synthetic visual instruction data~\cite{liu2023visual} with high-quality human-annotated multimodal dialogues. The second, LLaVA-RLHF~\cite{sun2023aligning}, leverages a reward model initialized from LLaVA-SFT-13B and applies the FACT-RLHF algorithm to the same data mixture. 

All evaluations are conducted using the same hyperparameters for contrastive decoding ($\alpha = 1$, $\beta = 0.1$), and each experiment is repeated three times with different random seeds to ensure robustness. The final scores reported are the averages over these runs. For benchmarks where VCD~\cite{leng2023mitigating} does not report generation parameters, we re-implement and report the results for a fair comparison. It is noteworthy that, at the time of our experimentation, the previously employed GPT-4 API (gpt-4-0314), as utilized in established benchmarks~\cite{sun2023aligning,liu2023improvedllava}, has been deprecated. Consequently, we adopt the updated GPT-4 API (gpt-4-0613) and present the re-evaluated performance in this paper. 

\begin{table}[t]
  \centering
  \caption{Performance of various debiasing strategies on MME. The notation ``VCD (u/n)'' indicates the application of the VCD decoding strategy with either the None (u) or Unk (n) Post-Hoc Debias methods.}
  \resizebox{\linewidth}{!}{%
  \begin{tabular}{@{}c cccccccc@{}}
  \toprule
  \multicolumn{9}{c}{LLaVA-v1.5-7B} \\ \midrule
  \multirow{2}{*}{\centering Subsets} & \multirow{2}{*}{\centering Naive} & \multicolumn{3}{c}{Post-Hoc Debias} & \multirow{2}{*}{\centering VCD\cite{leng2023mitigating}} & \multicolumn{3}{c}{Visual Debias Decoding} \\ 
  \cmidrule(lr){3-5} \cmidrule(lr){7-9}
   &  & None & Unk & Both &  & VDD-None & VDD-Unk & Both \\
  \midrule
  MME-OH & 313.33 & \textbf{335.00} & 328.33 & 303.33 & 303.33 & 316.67 & 318.33 & 328.33 \\
  MME-AH & 260.00 & \textbf{315.00} & 305.00 & 275.00 & 283.33 & 296.67 & 291.67 & 291.67 \\
  MME-RE & 317.86 & 302.50 & 299.29 & 282.50 & 312.14 & 279.64 & 337.50 & \textbf{346.07} \\
  Overall & 891.19 & 952.50 & 932.62 & 860.83 & 898.81 & 892.98 & 947.50 & \textbf{966.07} \\
  \toprule
  \multicolumn{9}{c}{LLaVA-v1.5-13B} \\ \midrule
  \multirow{2}{*}{\centering Subsets} & \multirow{2}{*}{\centering Naive} & \multirow{2}{*}{\centering VCD\cite{leng2023mitigating}} & \multirow{2}{*}{\centering Unk} & \multirow{2}{*}{\centering VDD-Unk} & \multicolumn{4}{c}{Mixed Strategy} \\
  \cmidrule(lr){6-9}
   &  &  &  &  & VCD (u) & VCD (n) & VDD-None (u) & VDD-None (n) \\
  \midrule
  MME-OH & 291.67 & 316.67 & 343.33 & \textbf{351.67} & 338.33 & 333.33 & 348.33 & 348.33 \\
  MME-AH & 280.00 & 253.33 & 320.00 & 308.33 & 318.33 & 318.33 & \textbf{325.00} & 320.00 \\
  MME-RE & 357.14 & 284.29 & 277.86 & \textbf{361.43} & 279.29 & 281.79 & 267.50 & 269.29 \\
  Overall & 928.81 & 854.29 & 941.19 & \textbf{1021.43} & 935.95 & 933.45 & 940.83 & 937.62 \\
  \bottomrule
  \end{tabular}%
  }
  \label{tab:mme_debias_ablation}
\end{table}

\begin{figure}[h]
  \centering
  \begin{subfigure}[t]{0.48\linewidth}
    \centering
    \includegraphics[width=\linewidth]{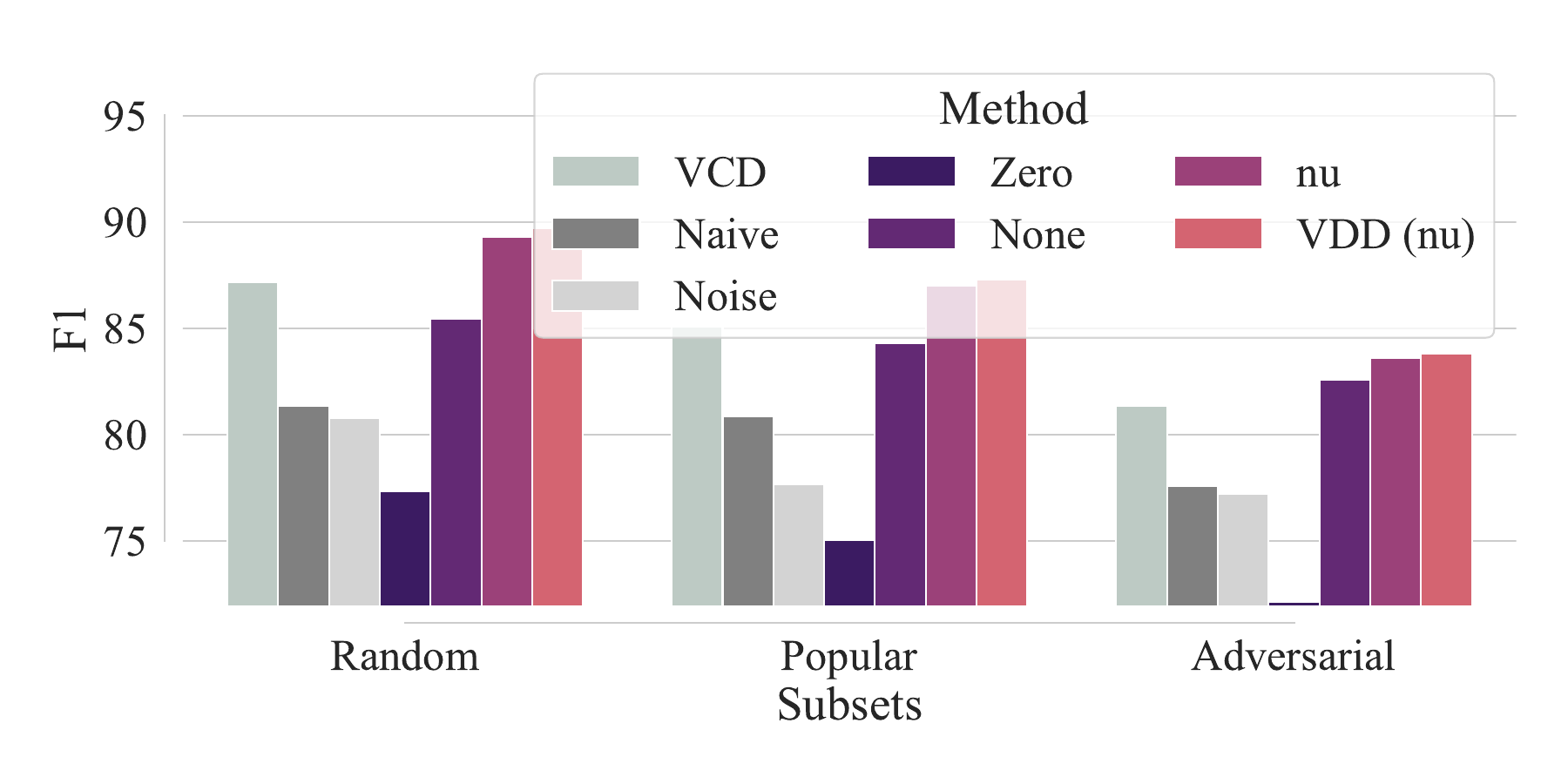}
    \vspace{-2em}
    \caption{\textbf{LLaVA-v1.5-7B}}
  \end{subfigure}%
  \hfill
  \begin{subfigure}[t]{0.48\linewidth}
    \centering
    \includegraphics[width=\linewidth]{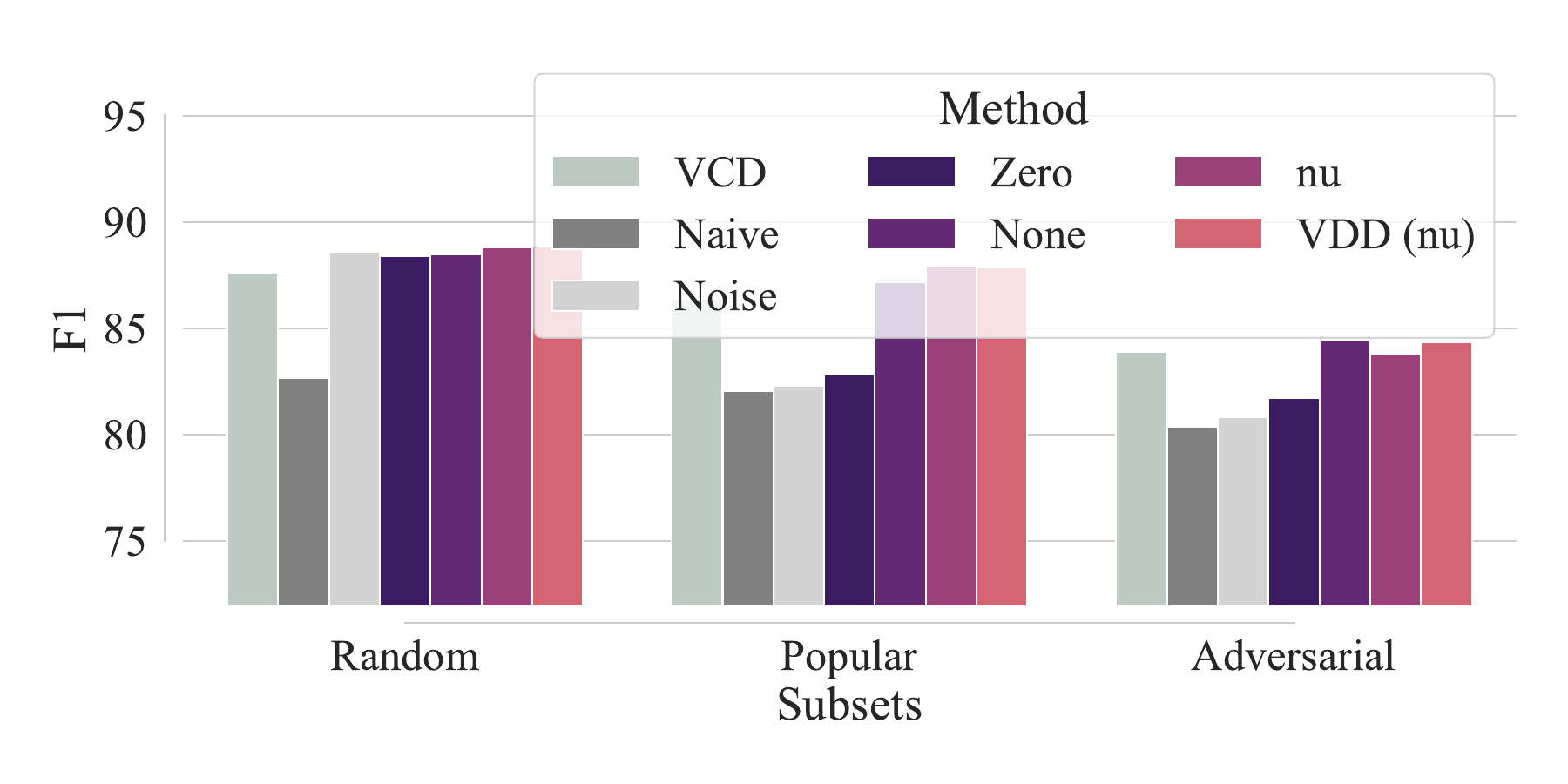}
    \vspace{-2em}
    \caption{\textbf{Qwen-VL}}
  \end{subfigure}
  \vspace{-1em}
  \caption{Comparison of various Post-Hoc Debias and Visual Debias Decoding methods on the POPE-COCO. ``nu'' denotes the application of both ``None'' and ``Unk'' for Post-Hoc Debias, and ``VDD-None'' is used by default.}
  \label{fig:cali_pope_ablation}
\end{figure}

\subsection{Impact of Post-Hoc Debias and Visual Debias Decoding Methods}

We conduct a comprehensive analysis by integrating multiple debiasing approaches—including Post-Hoc Debias and Visual Debias Decoding—to examine their effects on the truthfulness and reasoning capabilities of MLLMs. The results obtained from the MME dataset are presented in Table~\ref{tab:mme_debias_ablation}. Our key observations can be summarized as follows: (1) Post-Hoc Debias methods significantly improve model truthfulness by reducing hallucinations. Both the \textit{None} and \textit{Unk} strategies outperform the naive baseline, particularly in mitigating object-level and attribute-level hallucinations; (2) Visual Debias Decoding, such as VDD-None and VDD-Unk, achieve a better trade-off by maintaining strong hallucination control while delivering consistently higher reasoning performance; (3) Combining multiple Post-Hoc Debias strategies (Both) does not lead to notable improvements, whereas integrating debiased sampling methods yields promising gains; (4) Post-Hoc Debias and Visual Debias Decoding function independently. For example, VCD (u)—which combines the VCD sampling method with the Unk Post-Hoc Debias—achieves superior performance compared to VCD alone. Notably, VCD exhibits subpar performance with the LLaVA-13B model, likely due to its sensitivity to image noise levels. In contrast, our method relies on text rather than image, resulting in both higher efficiency and robustness. This trend is also observed on other benchmarks such as POPE (see Figure~\ref{fig:cali_pope_ablation}), where both our approaches consistently outperform naive and other baseline methods. Finally, we find that image-based debiasing techniques, such as Noise or Zero, fail to deliver consistent improvements. In our experiments, we observe a much larger variance and sensitivity to question type with such methods, further emphasizing the robustness of our image-free debiasing strategies

\subsection{Effect of Decoding Strategies}

\begin{table}[t]
  \centering
  \caption{Performance of different models with diverse decoding strategies.}
  \resizebox{\linewidth}{!}{%
  \begin{tabular}{ccccccc}
  \toprule
  \multicolumn{2}{c}{Setting} & \multirow{2}{*}{Default} & \multicolumn{3}{c}{Sampling} & \multirow{2}{*}{Overall} \\ \cmidrule{1-2} \cmidrule{4-6}
  Dataset & Model &  & Temp $\tau$ & Top-$p$ & Top-$k$ &   \\ \hline
  \multirow{4}{*}{\begin{tabular}[c]{@{}c@{}}POPE \\ MSCOCO\end{tabular}} & LLaVA-v1.5-7B & 79.5 & 84.0 & 84.0 & 84.0 & \textbf{84.1} \\
   & LLaVA-v1.5-13B & 77.2 & 84.2 & 84.2 & 84.1 & \textbf{84.2} \\
   & Qwen-VL & 81.8 & 83.6 & 83.7 & 83.3 & \textbf{83.7} \\
   & InstructBLIP & 78.9 & 79.8 & 84.5 & 79.9 & \textbf{84.5} \\ \cmidrule{2-7}
  \multirow{2}{*}{MMMU} & LLaVA-v1.5-7B & 34.4 & 43.4 & 40.3 & 40.8 & \textbf{44.7} \\
   & LLaVA-v1.5-13B & 36.3 & 43.2 & 43.4 & 43.1 & \textbf{45.8} \\ \cmidrule{2-7}
  \multirow{2}{*}{MME-OH} & LLaVA-v1.5-7B & 313.3 & 363.3 & 345.0 & 355.0 & \textbf{363.3} \\
   & LLaVA-v1.5-13B & 340.0 & 355.0 & 340.0 & 355.0 & \textbf{355.0} \\ \cmidrule{2-7}
  \multirow{2}{*}{MME-AH} & LLaVA-v1.5-7B & 260.0 & 323.3 & 301.7 & 325.0 & \textbf{330.0} \\
   & LLaVA-v1.5-13B & 303.3 & 321.7 & 311.7 & 321.7 & \textbf{321.7} \\ \cmidrule{2-7}
  \multirow{2}{*}{MME-RE} & LLaVA-v1.5-7B & 317.9 & 409.6 & 443.2 & 423.6 & \textbf{449.6} \\
   & LLaVA-v1.5-13B & 295.4 & 428.9 & 405.7 & 415.0 & \textbf{453.9} \\ \bottomrule
  \end{tabular}%
  }
  \label{tab:diverse_sampling_strategies}
\end{table}

\begin{table}[t]
  \centering
  \caption{Comparative analysis with baseline models.}
  \resizebox{\linewidth}{!}{%
  \begin{tabular}{ccccc}
  \toprule
  \rowcolor{gray!20}
  \textbf{Adversarial} & Accuracy & Precision & Recall & F1 \\
  LLaVA-1.5-7B\cite{leng2023mitigating} & 79.0 & 83.1 & 72.8 & 77.6 \\
  LLaVA-1.5-13B\cite{sun2023aligning} & 67.2 & - & - & 74.7 \\
  LLaVA-1.5-SFT-13B\cite{sun2023aligning} & 82.3 & - & - & 81.1 \\
  LLaVA-1.5-RLHF-13B\cite{sun2023aligning} & 82.3 & - & - & 80.5 \\
  LLaVA-1.5-7B (Ours) & 83.5 & 89.3 & 76.2 & 82.2 \\
  LLaVA-1.5-13B (Ours) & 84.3 & 91.6 & 75.6 & 82.8 \\ \midrule
  \rowcolor{gray!20}
  \textbf{Popular} & Accuracy & Precision & Recall & F1 \\
  LLaVA-1.5-7B\cite{leng2023mitigating} & 81.9 & 88.9 & 72.8 & 80.1 \\
  LLaVA-1.5-13B\cite{sun2023aligning} & 73.6 & - & - & 78.2 \\
  LLaVA-1.5-SFT-13B\cite{sun2023aligning} & 84.0 & - & - & 82.6 \\
  LLaVA-1.5-RLHF-13B\cite{sun2023aligning} & 83.9 & - & - & 81.8 \\
  LLaVA-1.5-7B (Ours) & 85.9 & 94.3 & 76.3 & 84.4 \\
  LLaVA-1.5-13B (Ours) & 86.1 & 95.3 & 75.9 & 84.5 \\ \bottomrule
  \end{tabular}%
  }
  \label{tab:comparative_analysis_with_baseline_models}
\end{table}

In Table~\ref{tab:diverse_sampling_strategies}, we comprehensively assess the performance of LLaVA-v1.5 across three distinct datasets—POPE-MSCOCO (average F1 across $3$ subsets), MME (total score across $3$ subsets), and MMMU (average Acc across $36$ subsets)—by employing various decoding strategies and settings. The ``Default'' column denotes the baseline performance using default sampling parameters, while the ``Sampling'' columns exhibit the results obtained through diverse decoding strategies, including temperature (Temp $\tau$), Top-$k$, and Top-$p$. The ``Overall'' column highlights the optimal performance achieved across all sampling strategies on all subsets. Notably, the application of the optimal sampling strategy leads to a substantial improvement in the overall model performance compared to default configurations. To underscore the generality of this phenomenon beyond LLaVA, we extend our evaluation to include Qwen-VL and InstructBLIP in the POPE-MSCOCO. The results consistently demonstrate performance enhancements with refined sampling strategies across existing MLLMs. In Table~\ref{tab:comparative_analysis_with_baseline_models}, we emphasize the superior performance of our proposed method, labeled as ``Ours'', in comparison to alternative strategies. Our method surpasses fine-tuned models on specific instruction datasets, highlighting the effectiveness of our refined sampling strategies. Moreover, the 7B model, employing these refined sampling strategies, outperforms the 13B model with different fine-tuning strategies, including both supervised fine-tuning and RLHF.

\begin{figure*}[h]
  \centering
  \begin{subfigure}[t]{0.32\linewidth}
    \centering
    \includegraphics[width=\linewidth]{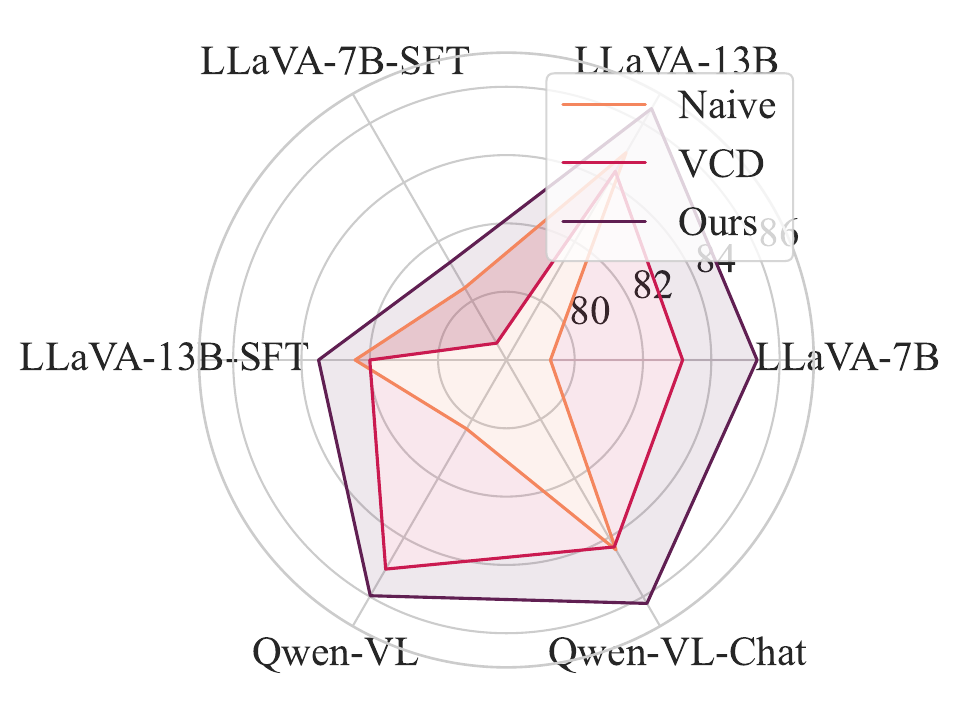}
    \caption{\textbf{POPE}}
  \end{subfigure}%
  \hfill
  \begin{subfigure}[t]{0.32\linewidth}
    \centering
    \includegraphics[width=\linewidth]{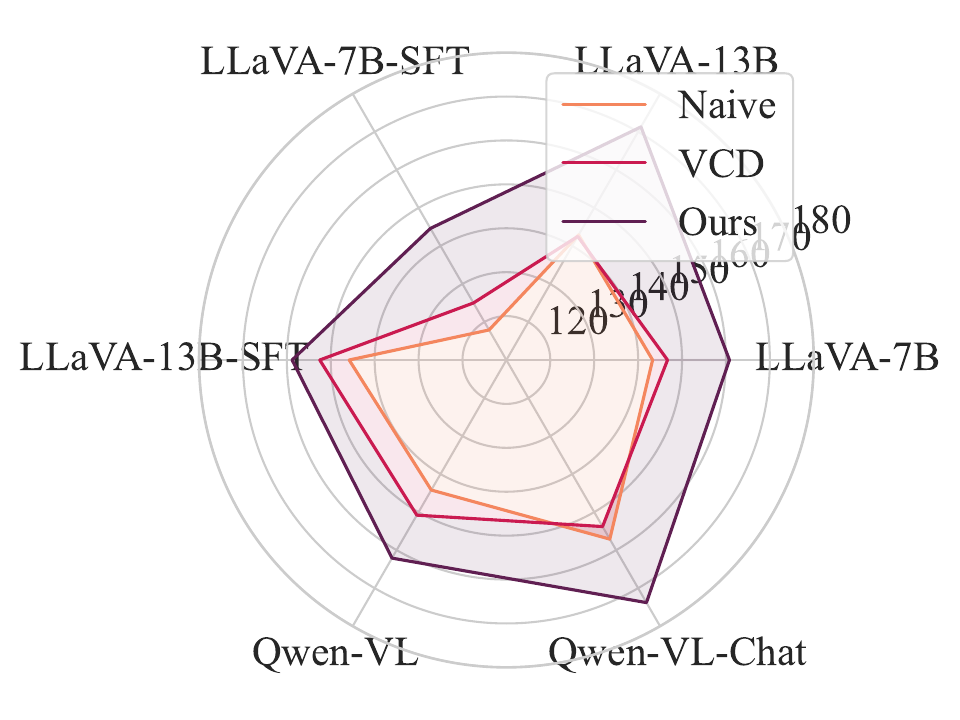}
    \caption{\textbf{MME-Hallucination}}
  \end{subfigure}%
  \hfill
  \begin{subfigure}[t]{0.32\linewidth}
    \centering
    \includegraphics[width=\linewidth]{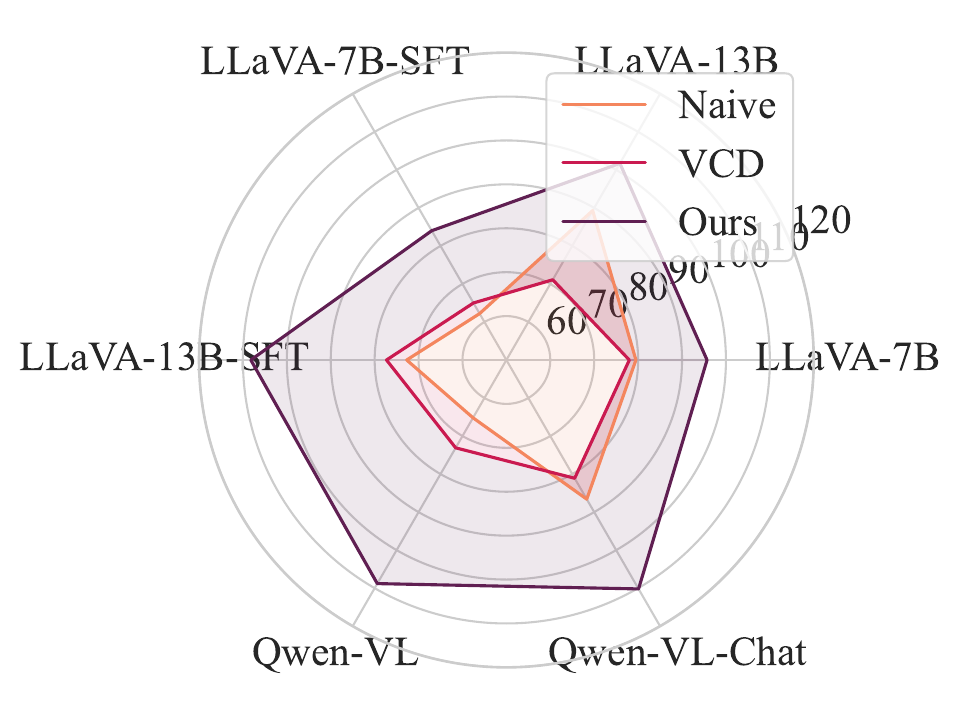}
    \caption{\textbf{MME-Reasoning}}
  \end{subfigure}%
  \vspace{-0.5em}
  \caption{Comparison with baselines across different benchmarks and varied backbone models.}
  \label{fig:main_com}
\end{figure*}

\subsection{Overall Performance}

\begin{figure}[h]
  \centering
  \begin{subfigure}{\linewidth}
    \centering
    \includegraphics[width=\linewidth]{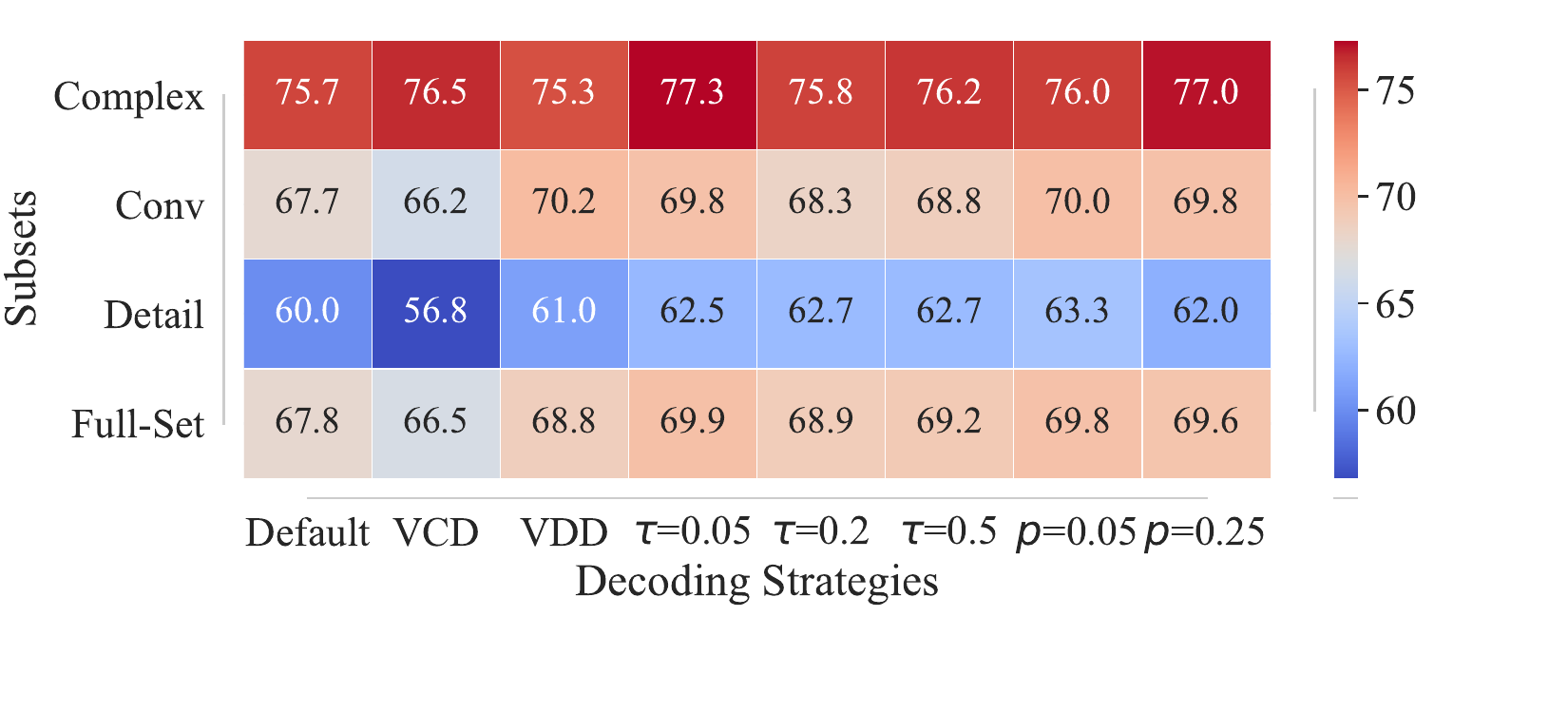}
    \vspace{-3.5em}
    \caption{\textbf{Absolute scores}}
  \end{subfigure}
  \hfill
  \begin{subfigure}{\linewidth}
    \centering
    \includegraphics[width=\linewidth]{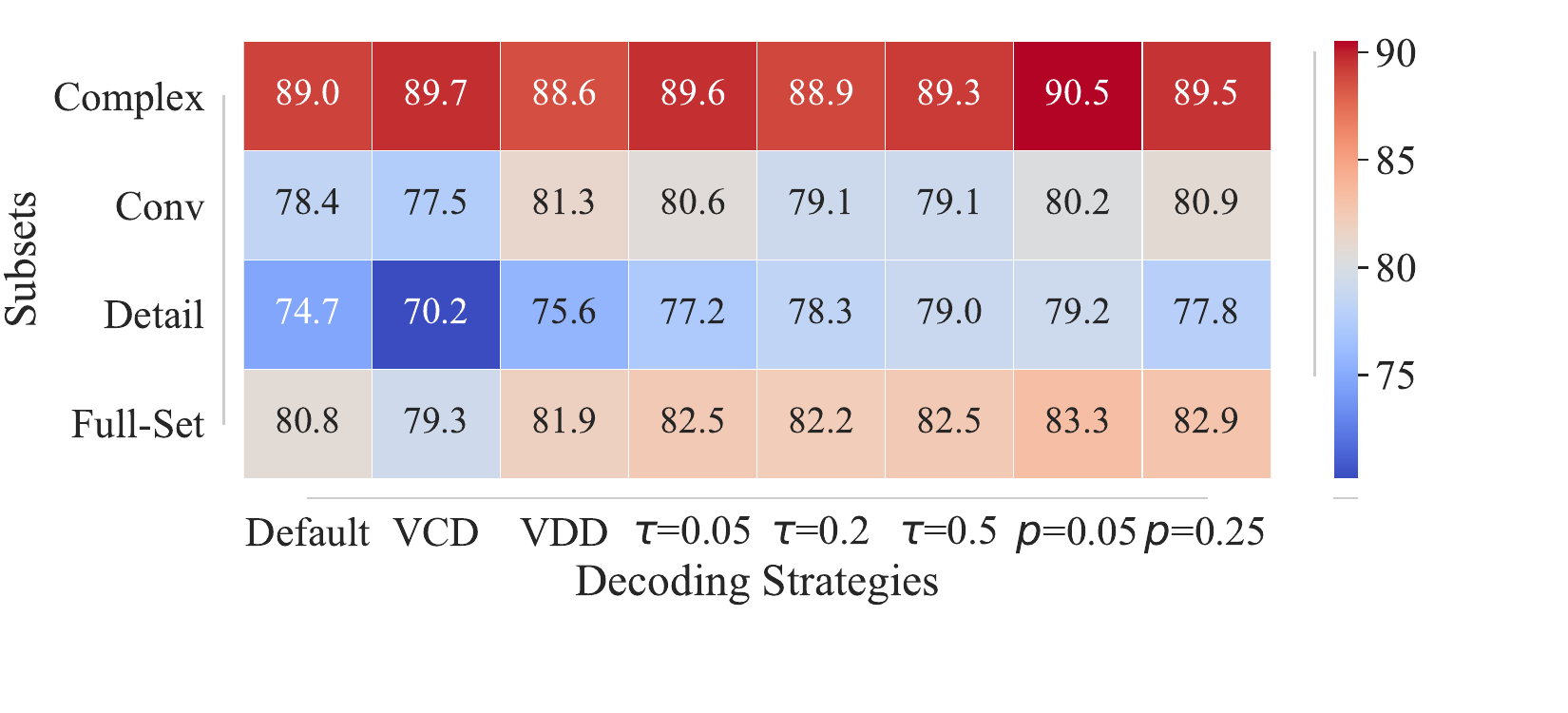}
    \vspace{-3.5em}
    \caption{\textbf{Relative scores}}
  \end{subfigure}
  \vspace{-2em}
  \caption{Automatic evaluation of LLaVA-v1.5-7B on LLaVA-Bench: GPT-4 compares model outputs with answers from GPT-4 (text-only) and assigns ratings in the range of $[1,10]$. We present both absolute scores (top), scaled up to 100, and relative scores (bottom) compared to GPT-4 (text-only).}
  \label{fig:main_llava_bench}
\end{figure}

Upon reaching the final iteration of our proposed method, we initially identify the optimal sampling configuration for all backbones. Subsequently, we apply debias sampling methods. For classification tasks, we employ Post-Hoc Debias using both None and Unk inputs concurrently. As shown in Figure~\ref{fig:main_com}, VCD fails to consistently improve performance across all baseline models, especially in reasoning tasks. In contrast, our proposed approach achieves consistently superior results across all evaluation dimensions and backbone models. Finally, the proposed method is assessed on the LLava-Bench, and the results presented in Figure~\ref{fig:main_llava_bench} demonstrate that, for open-ended generation tasks, VCD does not surpass the default decoding configuration consistently, although it occasionally outperforms in the complex subset. In contrast, our proposed VDD consistently exhibits improvements over the default decoding strategy. The right columns of the VDD section illustrate that, for generation tasks, model performance is consistently influenced by various decoding configurations. Unlike multi-choice QA tasks, which favor a low temperature ($\tau$), we observe that generation tasks benefit from a higher $\tau$ or a larger $p$ value.

\subsection{Findings and Analysis}

\begin{figure}[t]
    \centering
    \includegraphics[width=\linewidth]{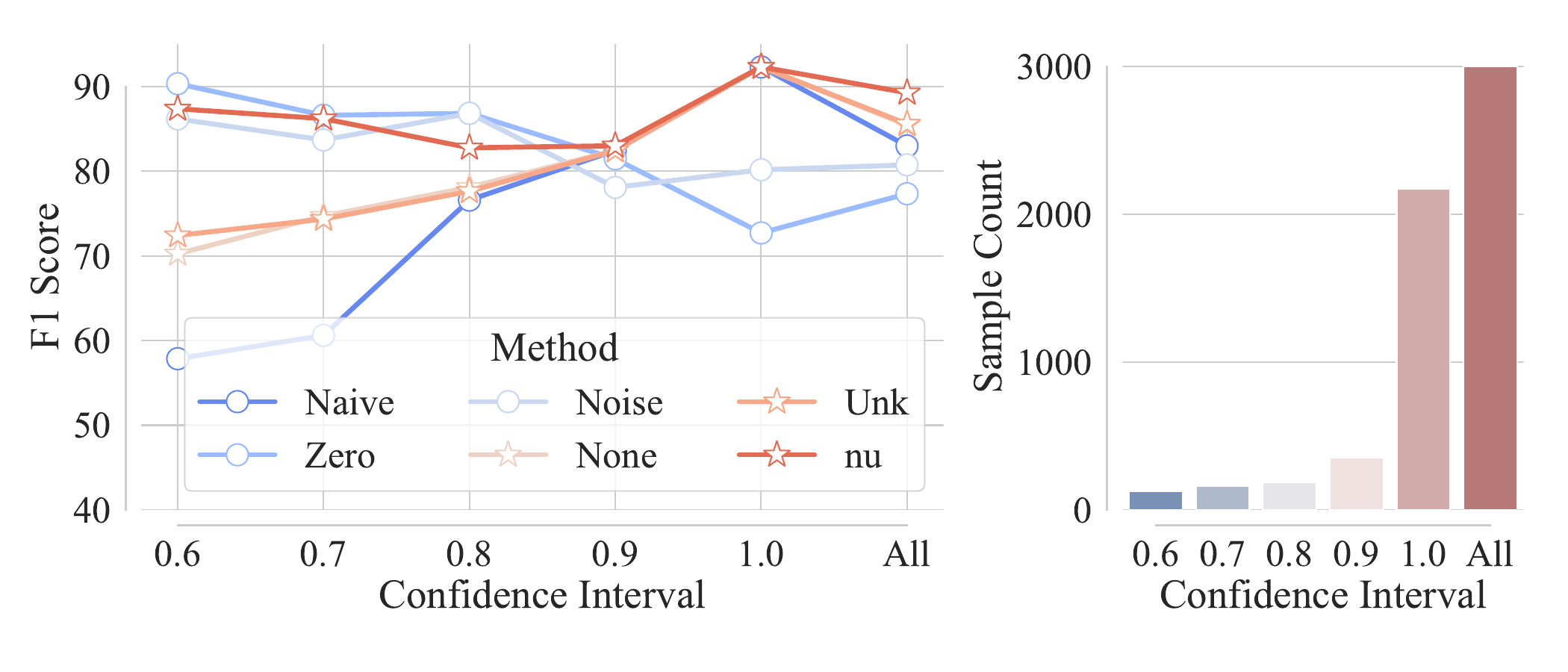}
    \vspace{-2em}
    \caption{Effectiveness of Post-hoc Debias across various confidence intervals.}
    \label{fig:confidence}
\end{figure}

\textbf{Calibration is particularly advantageous when a model lacks confidence in its predictions.} In Figure \ref{fig:confidence}, we analyze various Post-Hoc Debias methods. It is evident that the ``Naive'' approach performs poorly on instances where the model exhibits low confidence, which aligns with intuition. In contrast, our proposed Post-Hoc Debias methods yield significant performance improvements for such cases. As prediction confidence increases, the model's predictions become more reliable and accurate, resulting in diminishing performance gains. Even at high confidence levels, around $0.9$-$1.0$, our proposed methods achieve comparable results without detrimental effects. Consequently, the overall performance of our proposed method surpasses that of the naive approach. Notably, both image-based debiasing methods demonstrate superior performance on instances with low confidence scores, indicating aggressive adjustments to prediction results. However, such aggressive adjustments may lead to poorer predictions for high-confidence samples. Therefore, we default to using image-free debiasing methods, as they consistently yield improvements. 

\begin{figure}[t]
    \centering
\includegraphics[width=\linewidth]{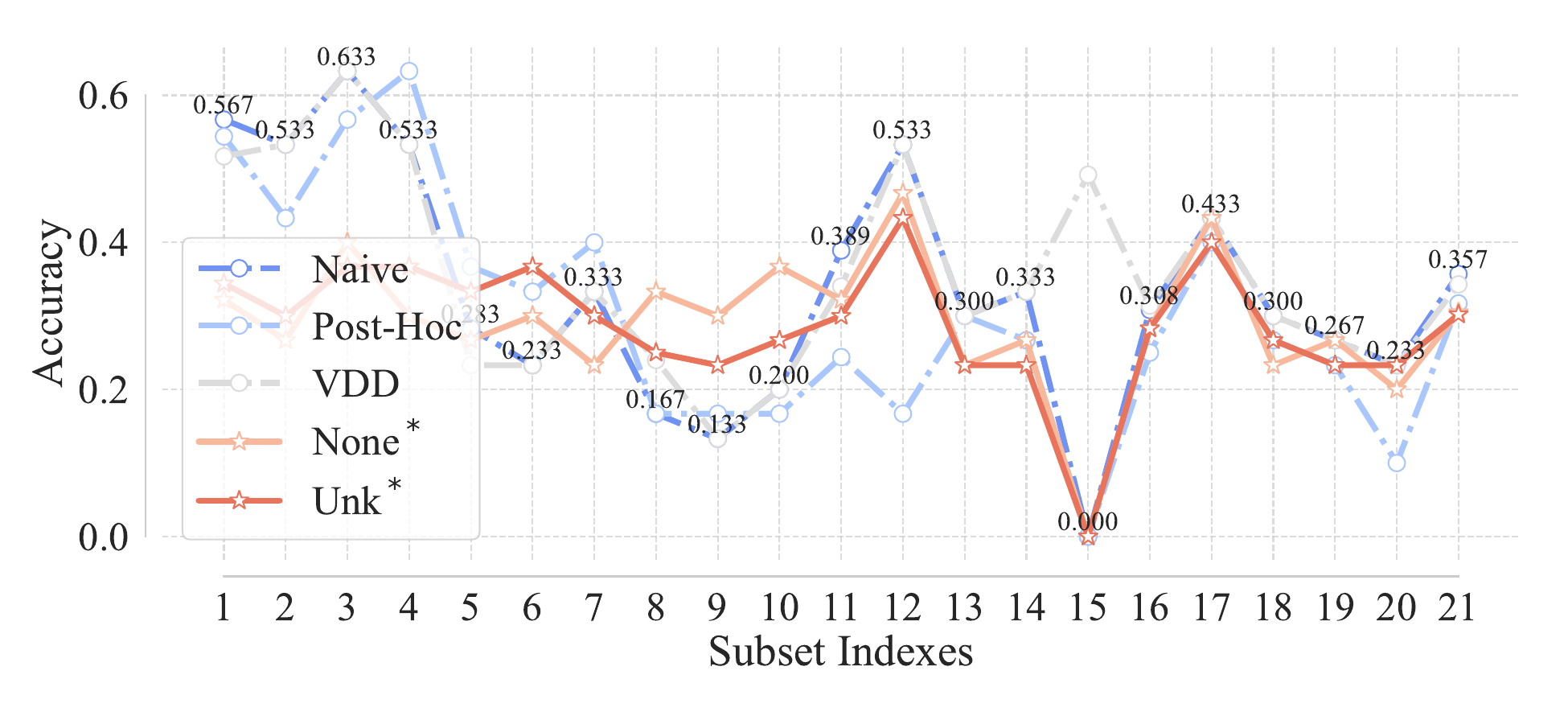}
\vspace{-0.5cm}
    \caption{Accuracy across subsets of LLaVA-1.5-7B on MMMU.}
    \label{fig:mmmu_wo_vision}
    \vspace{-0.4cm}
\end{figure}

\textbf{Navigating challenges in debiasing LLM-biased benchmarks.} As illustrated in Fig.~\ref{fig:mmmu_wo_vision}, we unravel the intricate interplay between debiasing methods and benchmarks influenced by LLM biases, with Naive's performance centered around textual information. The Post-Hoc Debias methods proposed exhibit inconsistent enhancements on the MMMU benchmark. Upon meticulous examination, when we entirely remove vision information by replacing the input image with an empty string (None$^*$) or the $<unk>$ identifier (Unk$^*$), the model maintains comparable or superior performance in specific subsets. In these scenarios, our Post-Hoc Debias methods, crafted to alleviate biases in MLLMs, fall short of yielding optimal results. Especially in instances where pure LLMs outshine, the application of Post-Hoc Debias methods may prove detrimental. It is crucial to emphasize that benchmarks tailored for MLLMs should prioritize reliance on input images rather than solely on textual content. Consequently, our methods can act as valuable indicators; when the proposed debiasing methods exhibit suboptimal performance, it signals that the benchmark may tilt more towards LLMs and may not be apt for evaluating MLLMs effectively. In such cases, the proposed sampling methods, VDD, also demonstrate inferiority compared to Naive, with an average accuracy of $34.3\%$ versus $35.7\%$.

\section{Conclusion}
We undertake a comprehensive exploration of language biases and challenges associated with Multimodal Large Language Models (MLLMs), particularly focusing on their interaction with underlying Large Language Models (LLMs). Our investigation exposes notable biases in MLLM-generated content, primarily influenced by language priors ingrained in LLMs rather than visual inputs. To address the biases, we introduce innovative debiasing strategies, including Post-Hoc Debias and Visual Debias Decoding. Our experiments demonstrate the effectiveness of these strategies in mitigating hallucination and enhancing reasoning capabilities in MLLMs. Post-Hoc Debias significantly improves model truthfulness, particularly when models lack confidence in their predictions. Moreover, Visual Debias Decoding strikes a balance by exhibiting competitive hallucination scores while consistently outperforming in reasoning tasks. The proposed strategies contribute to the reliability and applicability of MLLMs, addressing biases associated with language priors. Furthermore, our exploration into the impact of decoding configurations on MLLM performance reveals substantial improvements by refining sampling strategies. Optimal decoding configurations can unleash the full potential of existing MLLMs, often surpassing results obtained under default settings. This observation raises concerns regarding the fairness of current evaluation methods.


\clearpage

\begin{acks}
    This research is supported by the 2022A000300 Research Program of Intelligent Game and Decision Lab.
\end{acks}

\bibliographystyle{ACM-Reference-Format}
\balance
\bibliography{reference}

\end{document}